\documentclass[10pt, a4paper, twocolumn, teaser, showabstract]{naverlabseurope}

\usepackage{graphics}

\usepackage{subcaption}
\usepackage{wrapfig}
\usepackage{multicol}
\usepackage{tikz,tkz-kiviat,pgfplots}
\usepackage{cleveref}
\crefname{figure}{Fig.}{Figs.}

\newcommand{\ours}{DOCIR}
\newcommand{\mdp}[1]{\mathcal{#1}}

\graphicspath{{figures/}}

\title{\ours: Disentangled Object-Centric Image Representation for Robotic Manipulation}
\titlerunning{Disentangled Object-Centric Image Representation for Robotic Manipulation}

\correspondingauthor{[seungsu.kim, david.emukpere]@naverlabs.com \\ This work has been submitted to the IEEE for possible publication. Copyright may be transferred without notice, after which this version may no longer be accessible.}

\authors{
David Emukpere \authsep Romain Deffayet \authsep Bingbing Wu \authsep Romain Brégier Michael Niemaz \authsep Jean-Luc Meunier \authsep Denys Proux \authsep Jean-Michel Renders \authsep Seungsu Kim}

\affiliations{NAVER LABS Europe}
\contributions{}
\website{https://europe.naverlabs.com}
\websiteref{\href{https://europe.naverlabs.com}}

\teaserfig{
\includegraphics[width=0.95\linewidth]{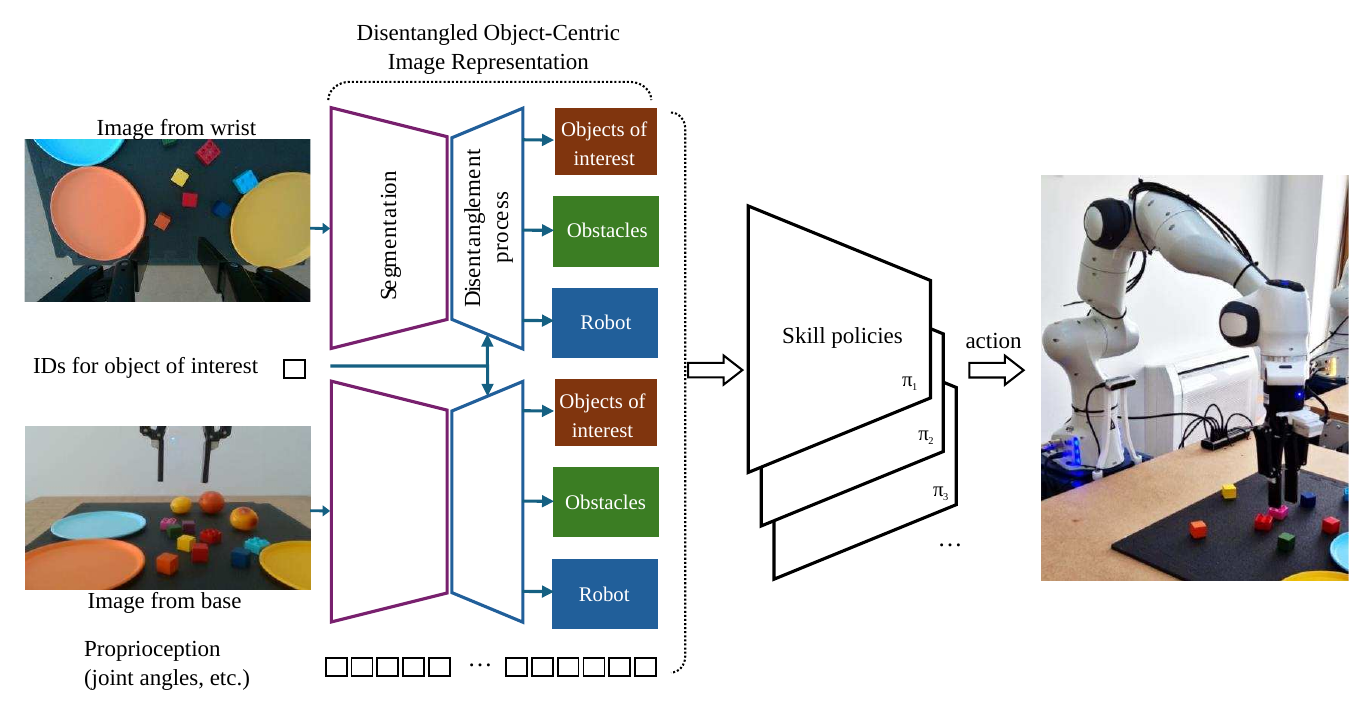}
\label{fig:docir}
}

\teasercaption{\textbf{DOCIR overview}. We introduce an object-centric disentanglement scheme for multi-view visual robotic manipulation to improve learning performance, efficiency, and generalization of robotic manipulation skills.}

\begin{abstract}

Learning robotic manipulation skills from vision is a promising approach for developing robotics applications that can generalize broadly to real-world scenarios. As such, many approaches to enable this vision have been explored with fruitful results. Particularly, object-centric representation methods have been shown to provide better inductive biases for skill learning, leading to improved performance and generalization. Nonetheless, we show that object-centric methods can struggle to learn simple manipulation skills in multi-object environments.

Thus, we propose DOCIR, an object-centric framework that introduces a disentangled representation for objects of interest, obstacles, and robot embodiment. We show that this approach leads to state-of-the-art performance for learning pick and place skills
from visual inputs in multi-object environments and generalizes 
at test time to changing objects of interest and distractors in the scene. Furthermore, we show its efficacy both in simulation and zero-shot transfer to the real world.

\end{abstract}

\begin{document}

\maketitle

\section{Introduction}

Manipulator robots are often deployed in carefully engineered setups. However, to solve a broader range of manipulation tasks, they must be flexible enough to accommodate diverse and sometimes unforeseen real-world scenarios. A popular approach towards this goal relies on learning vision-based manipulation policies, thereby discarding the need for careful engineering of the input space. Many studies have investigated how to directly use visual inputs to learn manipulation skills, whether to learn from demonstration~\cite{chi2023diffusionpolicy, chi2024diffusionpolicy, aloha, fu2024mobilealoha, VQBeT} or through reinforcement learning approaches~\cite{ Xiao2022MVP, wang2022vrl3, lfsbaseline}.

Despite promising performance on some manipulation tasks, the learned policies often fail to generalize to setups different from
those they have been trained on.
Using object-centric image representations as inductive bias appears as an effective approach to learn better policies with performance and generalization improvements~\cite{RobustnessImplicationsInObjectCentricLearning}.
Recent work~\cite{POCR} demonstrates how exploiting large-scale pretrained visual foundation models for object-centric robotic skill learning could lead to improved performance and better robustness to distractors and scene changes for various tasks~\cite{rlbench}.
Yet, this approach does not readily generalize to changing objects of interest at deployment time, and as we will show in the remainder, a naive extension of this method does not either.

\begin{figure*}[ht!]
    \centering
    \begin{subfigure}{0.49\textwidth}
    \includegraphics[width=\linewidth]{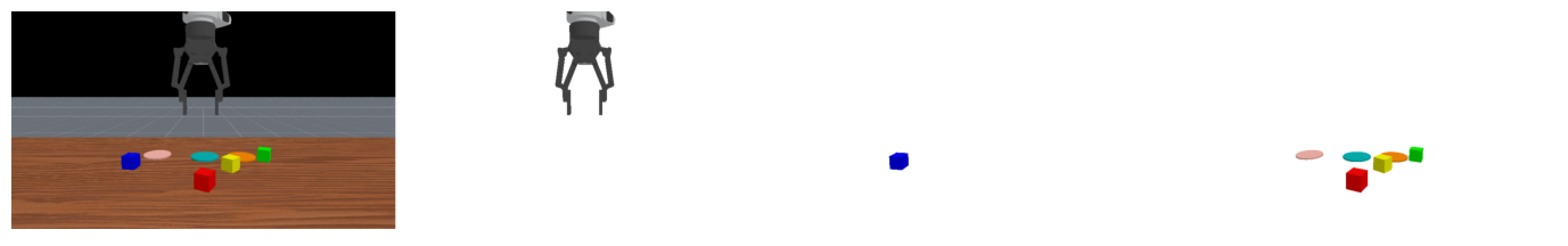}
    \includegraphics[width=\linewidth]{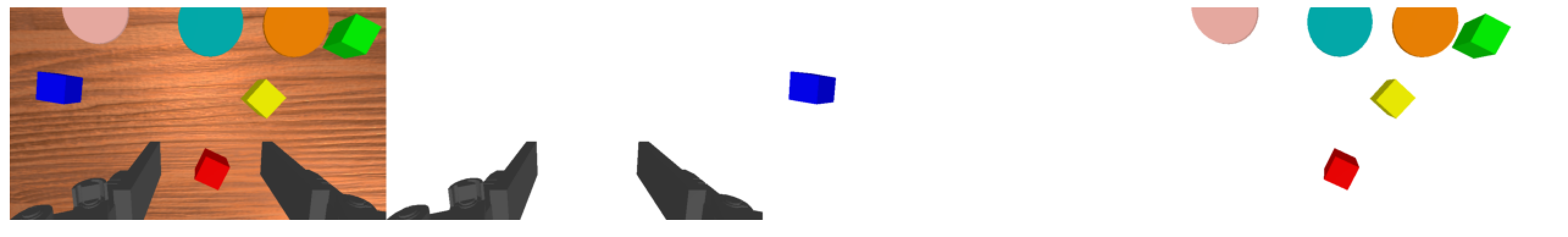}
    \caption{Simulated environment}
    \end{subfigure}
    \begin{subfigure}{0.49\textwidth}
    \includegraphics[width=\linewidth]{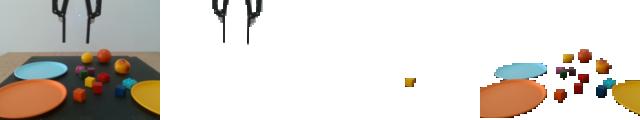}
    \includegraphics[width=\linewidth]{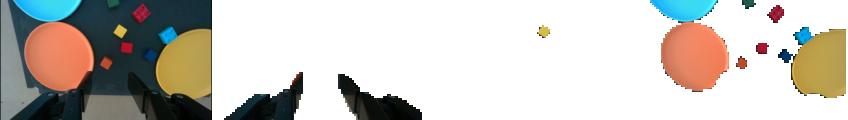}
    \caption{Real environment}
    \end{subfigure}
    \caption{\textbf{Example DOCIR segmentation}. Observation recovered by our disentanglement procedure 
    in simulation and real-world environments. In each environment, the two rows show the base and wrist camera views, respectively. Leftmost, is the full image, followed by the masked images for the robot, object of interest, and obstacles.}
    \label{fig:docir_segmentation_input}
\end{figure*}

Thus, we propose a framework that pushes object-centric representation learning further by introducing a disentangled representation for objects of interest, obstacles, and robot embodiment. We demonstrate that this disentangled approach leads to state-of-the-art performance for learning visual robotic policies in multi-object environments and generalizes more strongly at test time to changing objects of interest and scene layouts.

In summary, our main contributions are:
\begin{itemize}
    \item We introduce DOCIR, an object-centric framework that enables learning robust and generalizable robotic skills from visual inputs.
    \item We empirically validate the superiority of our method over
    state-of-the-art approaches for
    both scene-level and object-centric visual robotic skill learning through pick and place experiments.
    \item We demonstrate the robustness and generalization of DOCIR to distribution shifts at test time.
    \item We demonstrate the ability to transfer DOCIR-based policies to the real world.
\end{itemize}

\section{Approach}
\label{sec:approach}
We now describe our approach in detail: Disentangled Object-Centric Image Representation for Robotics (DOCIR). Our core idea is to obtain a structured object-centric image representation by processing images for each scene view to obtain three images corresponding to the following semantic groups:

\begin{itemize}
    \item \textbf{Robot}: This group contains all visible links of the robot in the image.
    \item \textbf{Objects}: This group contains the object(s) to be interacted with in the image, which depends on the skill to be executed and the IDs of the object(s) of interest.
    \item \textbf{Obstacles}: This group contains all other objects not to be interacted with in the image.
\end{itemize}

We refer to this procedure as our disentanglement process. Starting from a scene image, we first obtain a segmented image of constituent objects and the robot. In simulation, we have access to a ground-truth segmentation, while in the real-world we fine-tune a large vision model on our scene and objects that provides an equivalent segmented image\footnote{In~\Cref{subsec:real_world}, we further describe how we obtained this segmentation model for real-world experiments.}. Next we create binary masks from the segmented image as follows. For the robot mask, we set all pixels with IDs corresponding to robot links in the segmented image to 1 and the rest to 0. For the objects mask, we do the same with all pixels IDs corresponding to the object(s) of interest. Finally for the obstacles mask, we repeat this with all pixel IDs corresponding to other objects in the segmented image. We then apply these masks on the original scene image to obtain masked images corresponding to each group as illustrated in Fig.~\ref{fig:docir_segmentation_input}. Additionally, we concatenate the masked images with the binary masks used to generate them to obtain 4-channel images. This provides extra information enabling e.g.\ to disambiguate the segmentation of objects having a white color, which is used for masking. Furthermore, we note that as we adopt a multi-view approach with a base camera and robot wrist camera in the scene, we independently apply our disentanglement process to the images from both views.

Following our disentanglement process, we obtain image encodings for each group by processing the 4-channel images with a convolutional neural network (CNN). We use one shared CNN, $\mathrm{\phi_{view}}$, per camera view to obtain image encodings for each semantic group which are concatenated to form the camera view encoding $\Phi_{\mathrm{view}}$ i.e. 

\begin{align*}
    \Phi_{\mathrm{view}} = \left[ \mathrm{\phi_{view}(robot)}~||~ \mathrm{\phi_{view}(objects)}~||~\mathrm{\phi_{view}(obstacles)}\right].    
\end{align*}

\noindent where $\mathrm{view} \in \{ \mathrm{base}, \mathrm{wrist} \}$ are the cameras in the scene. Our final scene representation is therefore composed of two camera encodings $\{\Phi_{\mathrm{base}}, \Phi_{\mathrm{wrist}}\}$.

In our experiments, we evaluate the efficacy of DOCIR for robotic manipulation skill learning by using our object-centric disentanglement procedure while training a neural network policy using model-free reinforcement learning (RL)~\cite{SuttonAndBarto}. We represent each task as a Markov Decision Process~\cite{Puterman1994MarkovDP}  $\mdp{M} = \bigl \langle \mdp{S}, \mdp{A}, \mdp{P}, \mdp{R}\bigr \rangle$. $\mdp{S}$ is the state space consisting of the two images
from our multi-view cameras, along with robot proprioception variables (joint positions, joint velocities, end-effector pose, previous action, and boolean grasp status)\footnote{We make the common approximation that such observations are sufficient to guarantee the Markov property.}. $\mdp{A}$ is the action space with actions $a \in \mathbb{R}^{4}$ split into two parts: $ a_{\text{arm}} \in [-1, 1]^3$ is a 3D normalized end-effector displacement command, and ${a_{\text{gripper}}} \in \{-1, 1\}$ is a Boolean action to open or close the gripper. 
$\mdp{P}$ is the unknown transition function defining our environment dynamics, $\mdp{R}$ is the reward function containing skill-specific terms such as reaching, grasping, and obstacle avoidance. We train all policies using proximal policy optimization (PPO)~\cite{ppo}.

\section{Related Work}
\label{sec:related}
\subsection{Disentangled representation learning}
Disentangled representation learning ~\cite{Bengio_2013, higgins2018definition, Wang_2024} refers to machine learning techniques that extract latent representations corresponding to independent factors of variation in observable data. This approach is often motivated by its potential to enhance explainability and improve generalization in learned models.
In robotics, several studies have explored the advantages of disentangled representations. For example, ~\cite{Qian_2022} applies disentangled representation learning to encode object attributes—such as position, color, and shape—within a manipulation scene. In contrast, ~\cite{knowthyselftransferablevisual} proposes a method that decomposes the visual scene into agent and environment by leveraging known robot and camera parameters to generate a segmentation mask for the robot. This allows for learning a model-based policy that can be transferred across different robot embodiments given the robot parameters.
Unlike ~\cite{Qian_2022}, our approach does not focus on learning latent representations for individual objects. Instead, we decompose the original scene into semantic groups using segmentation masks. Similar to ~\cite{knowthyselftransferablevisual}, our disentangled representation distinguishes the robot from the environment in a visual scene. However, we extend this decomposition by further segmenting the environment into objects of interest and obstacles based on the task requirements. We demonstrate that this structured decomposition significantly improves learning in complex multi-object robotic manipulation tasks, where previous methods struggle, and enables better generalization to novel objects during deployment.

\subsection{Object-centric representation learning for robotics}
Learning object-centric representations for robotics has been extensively studied~\cite{Devin2017DeepOR, locatello2020object, SORNET, M2T2, VIOLA}. In general, it has been shown that compared to scene-level flat representations, object-centric representations yield improved performance, generalization, and robustness benefits~\cite{RobustnessImplicationsInObjectCentricLearning}. A critical recent development looks into composing pre-trained vision models to derive object-centric representations~\cite{POCR}. Our work differs from this and previous approaches by further incorporating a disentanglement to the visual scene to obtain a more structured object-centric representation that distinguishes object(s) of interest from surrounding objects. This design provides a more substantial inductive bias to the learning algorithm and enables strong generalization at test time to unseen objects of interest. Additionally, our approach does not require setting a fixed number of object slots to be discovered beforehand like POCR~\cite{POCR}, allowing a more scalable generalization to different numbers of objects in the scene.

\medskip 

\section{Experiments}
\label{sec:experiments}
We aim to answer the following questions empirically:
\begin{itemize}
    \item Does DOCIR improve the performance and learning efficiency of robotic policies learned from visual inputs?
    \item Does DOCIR improve the robustness and generalization of visual robotic policies?
    \item Can skills trained with DOCIR in simulation transfer to the real world?
\end{itemize}

\subsection{Setup}
\begin{figure}[h!]
    \centering
    \includegraphics[height=2cm, width=0.3\linewidth]{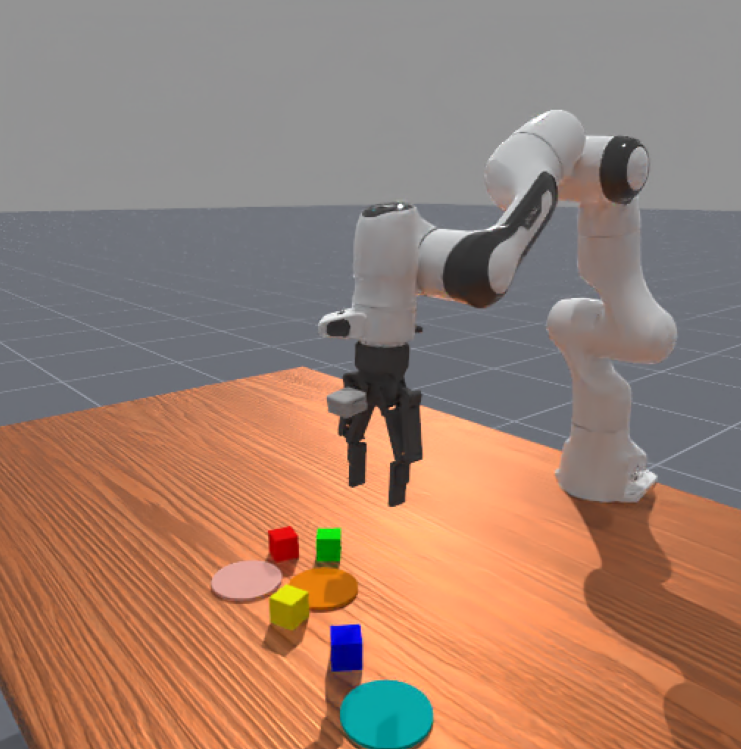}
    \includegraphics[height=2cm, width=0.33\linewidth]{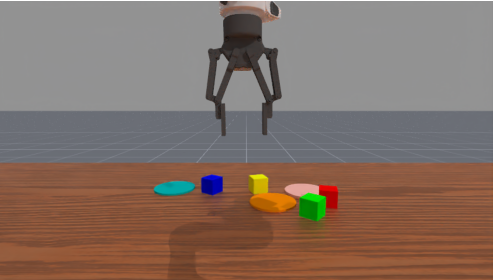}
    \includegraphics[height=2cm,width=0.33\linewidth]{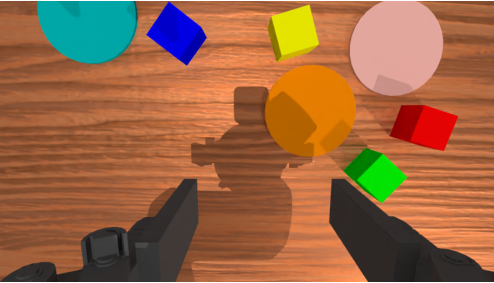}
    \caption{\textbf{Multi-object simulation environment}. (Left-to-right): full scene view, base camera view, wrist camera view.}
    \label{fig:simulation_env}
\end{figure}

We build a multi-object environment for our simulation experiments illustrated in Fig.~\ref{fig:simulation_env}, using ManiSkill~\cite{ManiskillDoc} as our simulation platform. The environment consists of a Franka Panda arm equipped with a Robotiq gripper, different colored cubes and plates on a table. 

In this environment, we train two robotic manipulation skills sequentially, namely: 
\begin{itemize}
    \item \textbf{Pick}: the goal is to pick one of the cubes without displacing other objects in the scene.
    \item \textbf{Place}: the goal is to place a picked cube on a target cube or plate in the scene without displacing the target and other objects.
\end{itemize}

Our initial states for the Place skill are obtained from successful terminal states of the Pick skill. Also, we use the same set of initial states, obtained from the best performing Pick skill for all methods to ensure a fair comparison.

\subsection{Baselines}
We consider the following baselines for our experiments:
\begin{itemize}
    \item \textbf{Flat scene representation}: This baseline is a typical approach that uses the images directly as input to the visual policy module. Our choice for this baseline is inspired by recent work~\cite{lfsbaseline} showing a learning-from-scratch baseline competitive with large-scale pretraining-based methods.
    
    \item \textbf{Object Centric representation}: For this baseline, We choose to implement a POCR~\cite{POCR}-like method, as POCR is a recent object-centric approach that was shown to deliver state-of-the-art performance. The primary difference is that we do not use pre-trained vision modules but provide explicit segmentation masks and learn the masked image encoder from scratch, to ensure meaningful comparisons with DOCIR. Unlike the disentangled nature of DOCIR, this baseline utilizes individual masked images for each object. Since we do not employ pretrained models here, we call this baseline OCR for Object Centric representation.
\end{itemize}

Both of these baselines lack a way to identify the object of interest in our multi-object environments. Therefore, to allow a fair comparison with our disentangled approach, we equip them with a learned object ID embedding that indicates the object of interest in our multi-object environments.

The skill policies employed in this study are multi-layer perceptrons (MLPs). To ensure a fair comparison, these baseline models are combined with the same architectural configuration of our skill policies. It's important to note that the self-attention mechanism utilized in the real-world robot experiments described in ~\cite{POCR} is not incorporated into our policy architecture, aligning with the approach taken in the simulated experiments of ~\cite{POCR}. 

\subsection{Tasks}
\label{subsec:tasks}
We train each of the two skills (Pick and Place) on two variants of the multi-object manipulation task:
\begin{itemize}
    \item \textbf{Fixed target}: Here, the scene consists of multiple objects, but the skills only need to interact with a unique object of interest in all episodes.
    
    \item \textbf{Varying target}: The scene also consists of multiple objects, but the skills must interact with a different object of interest at every new episode.
    
\end{itemize}

For both task variants, we also vary the complexity of the scene by modifying the numbers of objects on the table. Specifically, we use the following settings for the objects on the tabletop: \begin{itemize} 
\item 3 objects: 2 cubes, 1 plate;
\item 5 objects: 3 cubes, 2 plates; 
\item 7 objects: 4 cubes, 3 plates; 
\item 9 objects: 5 cubes, 4 plates. \end{itemize}

\subsection{Results}

\begin{figure}[t!]
\centering
\begin{subfigure}{\linewidth}
        \includegraphics[width=\textwidth]{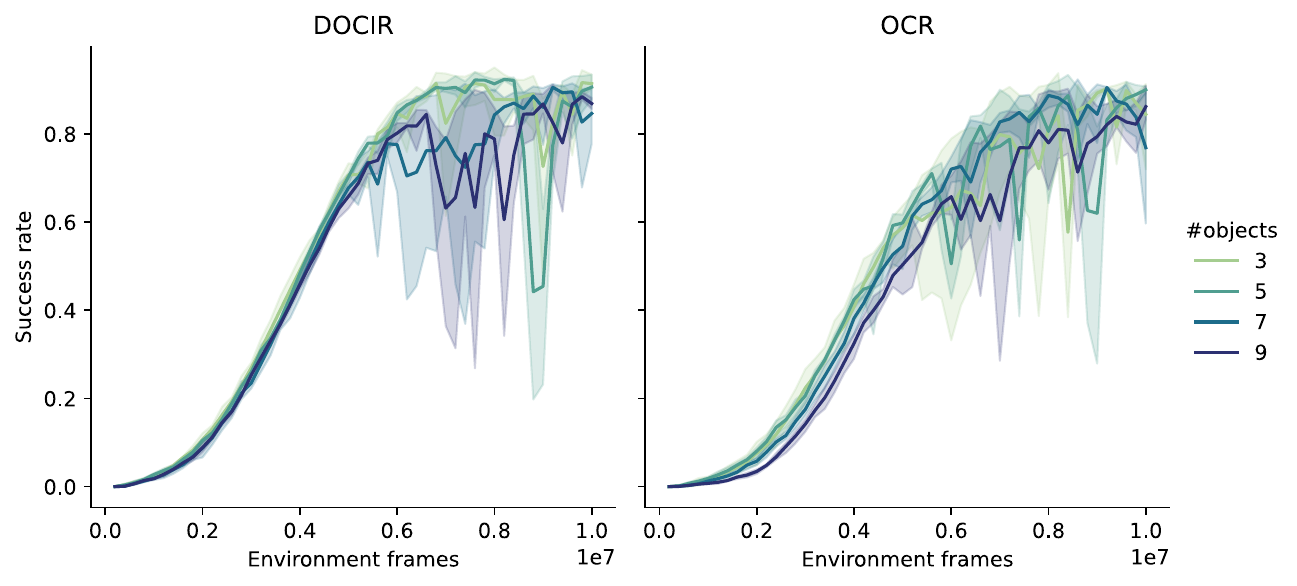}
\caption{Pick}
\end{subfigure}
\begin{subfigure}{\linewidth}
    \includegraphics[width=\textwidth]{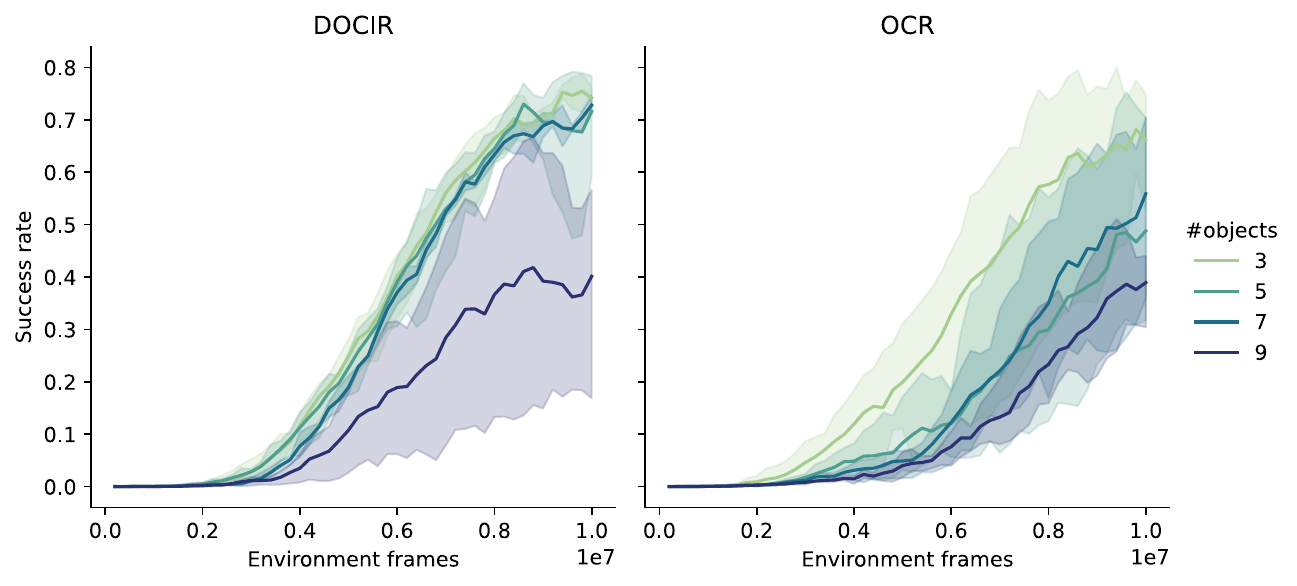}
    \caption{Place}
\end{subfigure}

\caption{\textbf{Fixed target skill learning curves.} We report a rolling average of training performance, aggregated over $3$ seeded runs. In the fixed target environment, the object of interest is the same across all episodes.}
\label{fig:fixed_target_learning_curves}
\end{figure}

\begin{figure}[t!]
\centering
\begin{subfigure}{\linewidth}
        \includegraphics[width=\textwidth]{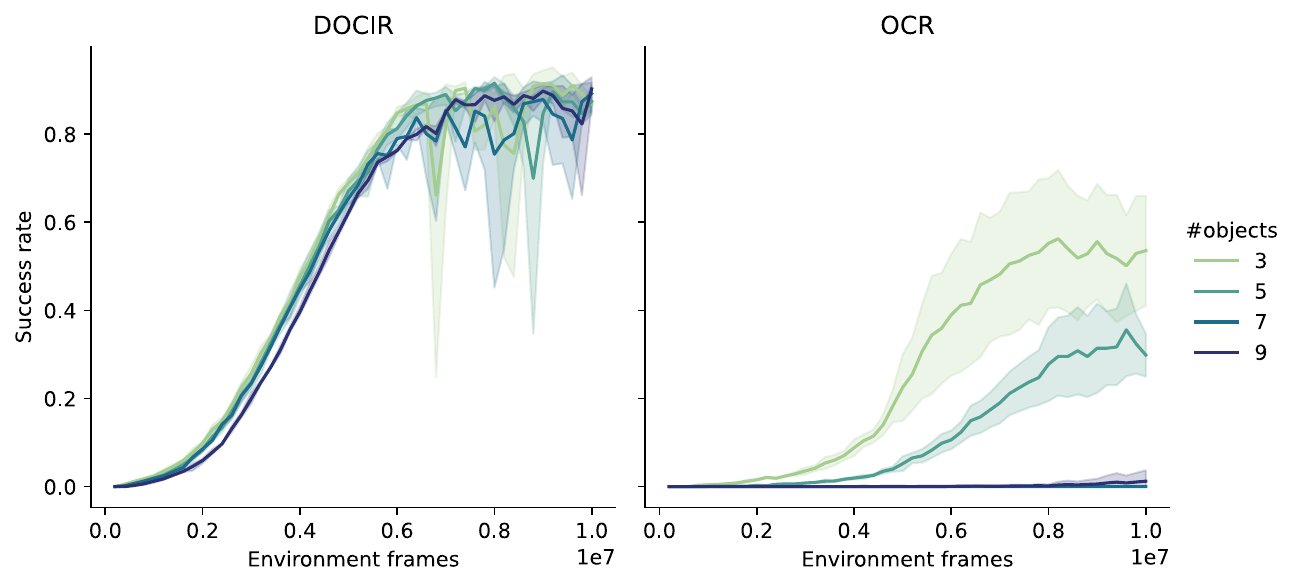}
\caption{Pick}
\end{subfigure}
\begin{subfigure}{\linewidth}
    \includegraphics[width=\textwidth]{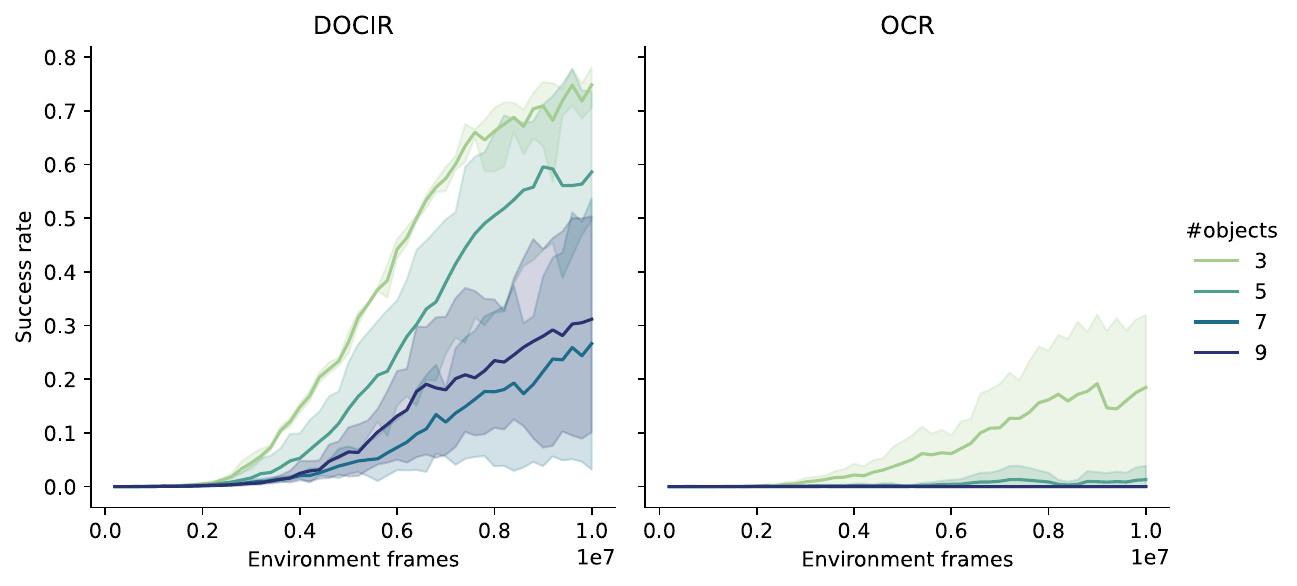}
    \caption{Place}
\end{subfigure}

\caption{\textbf{Varying target skill learning curves.} We report a rolling average of training performance, aggregated over $3$ seeded runs. In the varying target environment, the object of interest is different in every new episode.}
\label{fig:varying_target_learning_curves}
\end{figure}

\begin{figure}[t!]
\centering
\begin{subfigure}{\linewidth}
        \includegraphics[width=\linewidth]{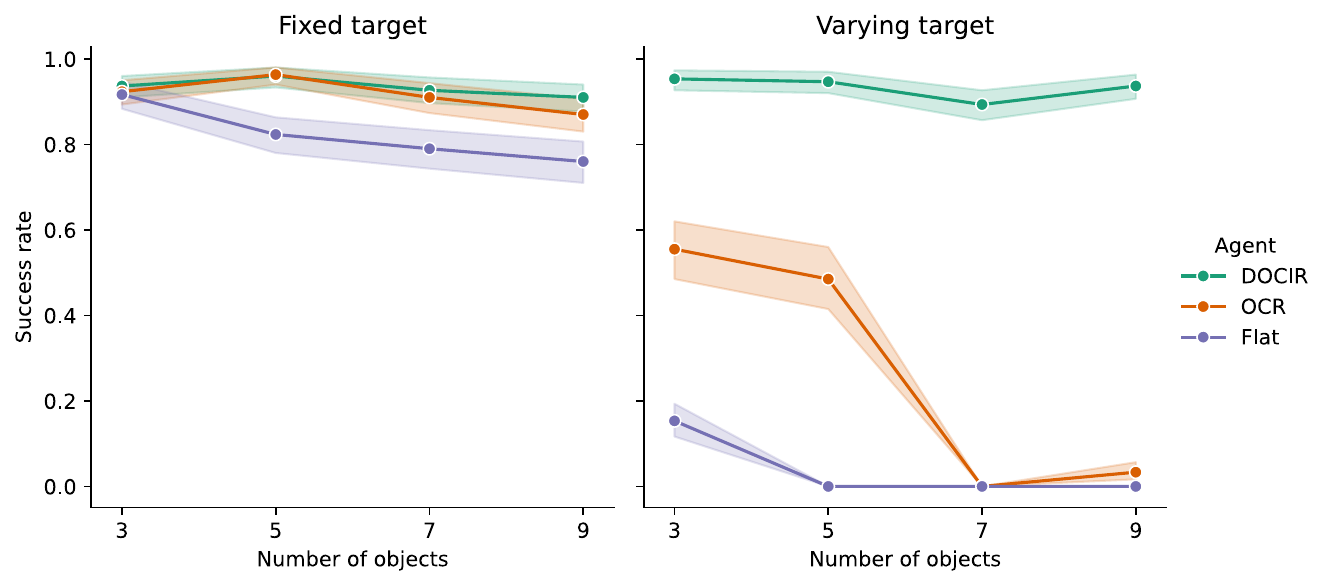}
\caption{Pick}
\end{subfigure}
\begin{subfigure}{\linewidth}
    \includegraphics[width=\linewidth]{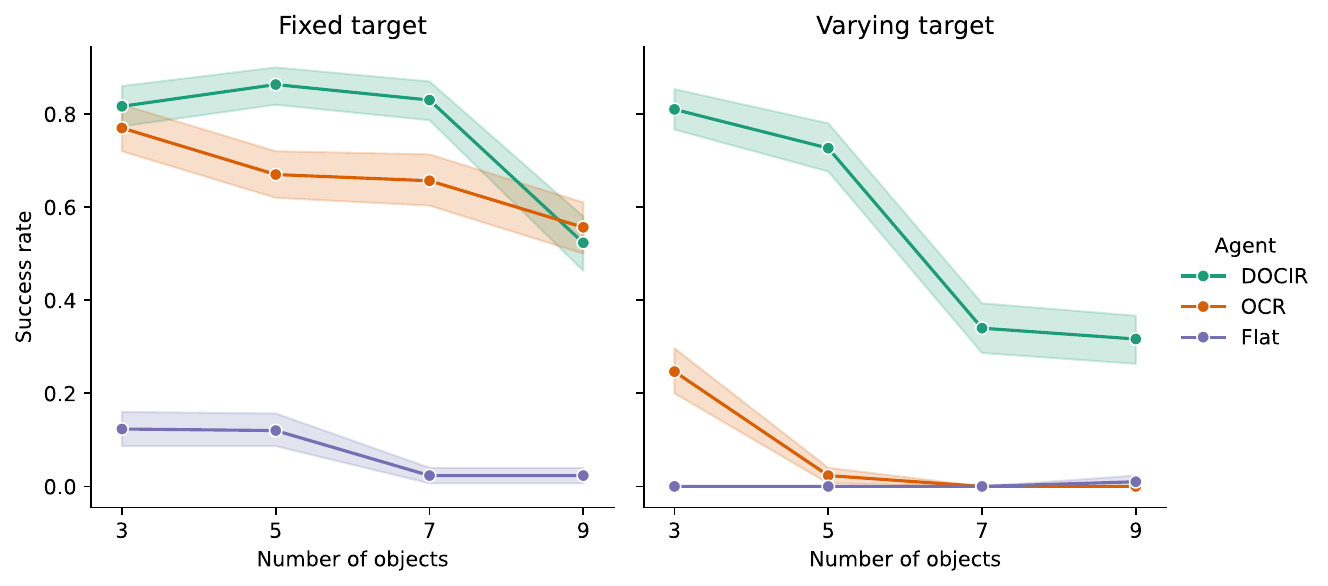}
    \caption{Place}
\end{subfigure}
    \centering
    
    \caption{\textbf{Evaluation results.} All agents are trained for $10$ million environment steps. We report here the average success rate of 3 seeded runs, by evaluating the best checkpoint of each run on $100$ test episodes with policy stochasticity turned off.}
    \label{fig:success_trend}
\end{figure}

\cref{fig:success_trend} provides an overview of the evaluation results for all methods on all environments, while \cref{fig:fixed_target_learning_curves} and \cref{fig:varying_target_learning_curves} show the learning curves of DOCIR and OCR for both skills in the \textbf{Fixed target} and \textbf{Varying target} variants respectively. Finally, we also report the full evaluation results in~\Cref{tab:skill_results}.

For the \textbf{Fixed target} tasks, we observe similar satisfactory performance of both object-centric methods (DOCIR and OCR) on the \textit{Pick} skill, and a slight edge for DOCIR on \textit{Place}. Even in this simple setup, the flat baseline struggles, as it reaches slightly lower performance on \textit{Pick} and does not learn to solve \textit{Place} at all. This observation is consistent with prior works showing the advantage of object-centric representations. 

On \textbf{Varying target} variants, DOCIR clearly outperforms OCR, while the flat baseline fails to learn even to \textit{Pick}. DOCIR's performance on \textit{Pick} remains very robust to the increasing complexity introduced by more objects in the scene, and while its performance on \textit{Place} is affected by the number of objects, it is the only method able to place on the correct target when many objects are present in the scene. In contrast, OCR's performance collapses as we increase the scene complexity.

\begin{table}[t!]
\resizebox{0.5\textwidth}{!}{
\begin{tabular}{lcccc|cccc}
\toprule
 & \multicolumn{4}{c}{Pick} & \multicolumn{4}{c}{Place} \\
\cmidrule(lr){2-5} \cmidrule(lr){6-9}
 (Cubes, Plates)& (2, 1) & (3, 2) & (4, 3 )& (5, 4) & (2, 1) & (3, 2) & (4, 3) & (5, 4) \\
\midrule
\multicolumn{1}{l}{\textbf{Fixed target}} \\
\midrule
DOCIR & \textbf{0.94} & \textbf{0.96} & \textbf{0.93} & \textbf{0.91} & \textbf{0.82} & \textbf{0.86} & \textbf{0.83} & 0.52 \\
OCR & 0.92 & \textbf{0.96} & 0.91 & 0.87 & 0.77 & 0.67 & 0.66 & \textbf{0.56} \\
Flat & 0.92 & 0.82 & 0.79 & 0.76 & 0.12 & 0.12 & 0.02 & 0.02 \\
\midrule
\multicolumn{1}{l}{\textbf{Varying target}} \\
\midrule
DOCIR & \textbf{0.95} & \textbf{0.95} & \textbf{0.90} & \textbf{0.94} & \textbf{0.81} & \textbf{0.73} & \textbf{0.34} & \textbf{0.32}\\
OCR & 0.56 & 0.49 & 0.00 & 0.03 & 0.25 & 0.02 & 0.00 & 0.00\\
Flat & 0.15 & 0.00 & 0.00 & 0.00 & 0.00 & 0.00 & 0.00 & 0.01 \\
\bottomrule
\end{tabular}
}
\caption{
    \textbf{Skill learning success rates}. We report here the average success rate of 3 seeded runs, by evaluating the best checkpoint of each run on 100 test episodes with policy stochasticity turned off.
    }
    \label{tab:skill_results}
\end{table}

\subsection{Generalization}
In this section, we evaluate the out-of-distribution (OOD) generalization performance of DOCIR-based policies. We consider two types of OOD settings:

\begin{itemize}
    \item \textbf{Unseen colors}: This setting involves changing the colors of the cubes and plates to colors unseen during training.

    \item \textbf{Distracting obstacles}: This setting includes extra objects to distract and increase scene clutter.
\end{itemize}

 For each setting, we evaluate a \textit{Pick} policy trained in the \{ 3 cubes, 2 plates \} training setup described in~\Cref{subsec:tasks}. Our results in~\Cref{tab:ood_success} show that DOCIR-based policies are robust to test-time distribution shifts across all settings considered.

\begin{table}[h!]
\centering
\begin{tabular}{lcc}
\toprule
Setting & Success rate\\
\midrule
In-distribution & 0.95 \\
Unseen colours & 0.86 \\
Distracting obstacles & 0.92 \\
\bottomrule
\end{tabular}
\caption{\textbf{Out-of-distribution success rates}. We compare in-distribution success rates with the out-of-distribution counterparts. Our results show that DOCIR generalizes adequately to OOD settings.}
\label{tab:ood_success}
\end{table}

\subsection{Ablations}
\begin{figure}[h!]
\centering
    \includegraphics[width=0.7\linewidth]{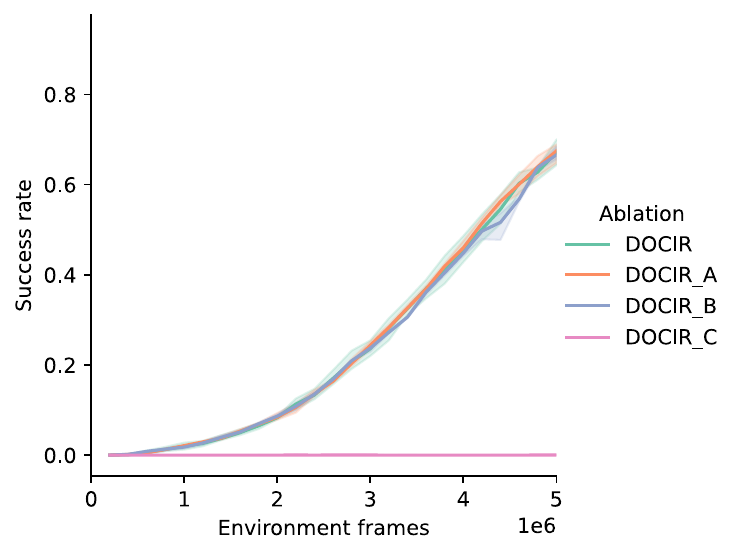}
    \caption{\textbf{Ablations}: We consider several variants of DOCIR with only two groups. Overall, our three-group separation design choice in DOCIR is similar to these variants.}
    \label{fig:docir_ablations}
\end{figure}
We ablate the choice of separation in our design by considering the following two-group variants that merge two out of the three classes into one, namely:
 \begin{itemize}
     \item \textbf{Ablation A}: Here, we combine the robot and objects in a mask separate from the obstacles mask.
     \item \textbf{Ablation B}: Here, we combine the robot and obstacles in a mask separate from the objects mask.
     \item \textbf{Ablation C}: Here, we combine obstacles and objects in a mask separate from the robot mask.
 \end{itemize}

We experiment with these variants to learn the \textit{Pick} skill in the \{ 3 cubes, 2 plates \} setting and visualize the resulting learning curves in~\cref{fig:docir_ablations}. Ablation A and Ablation B show a strong performance similar to DOCIR, while Ablation C fails. This result indicates that the most critical design decision to boost performance is separating objects of interest and obstacles in distinct groups. Although Ablations A and B perform on par with DOCIR, we believe our choice of three groups is more semantically meaningful and that having the robot in a standalone group could enable more effortless cross-embodiment transfer of skills.

\subsection{Real world experiments}
\label{subsec:real_world}
\begin{figure}[h!]
    \centering
    \includegraphics[width=.7\linewidth]{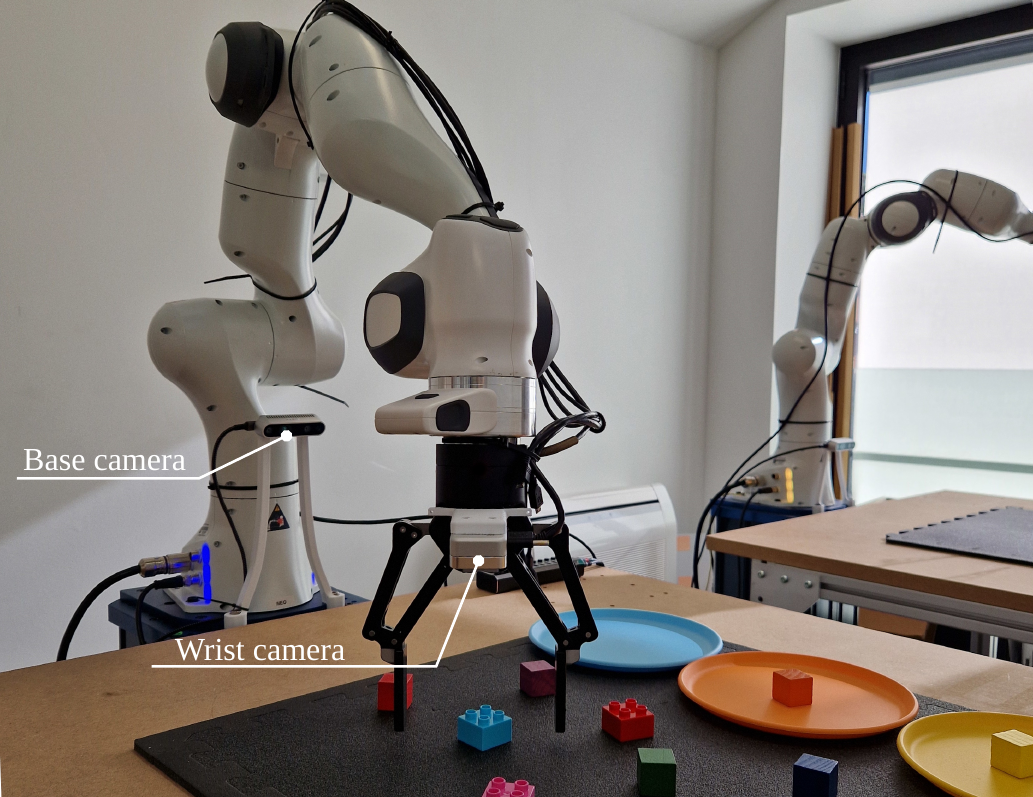}
    \caption{\textbf{Real world multi-object environment}. We deploy DOCIR-based policies learned in simulation in our real-world environment.}
    \label{fig:real_env}
\end{figure}

We validate DOCIR policies from the previous section in real-world manipulation tasks. This is challenging due to noisy observations from different sensors (e.g., cameras, sensors, etc.) and the different robot control systems between simulation and reality.

We use a 7-DOF Franka panda arm equipped with a two-finger gripper (Robotiq 2F-140 ) as shown in \cref{fig:real_env}. The robot is also equipped with two different cameras. The base camera (RealSense D435) is attached near the robot's base to capture a global view of the scene. The wrist camera (RealSense D405) is attached to the wrist of the robot to capture a closer view for manipulation. Note that we use only RGB information in our experiments. 

To obtain segmented images in the real-world, we fine-tune a ViT-small DINOv2~\cite{oquab2023dinov2} model on a small dataset of 826 wrist camera and 962 base camera images to output segmentation masks for objects and the robot in our scene. We obtained this fine-tuning dataset by first collecting 8 video sequences of predefined robot motions in our scenes. Subsequently, we utilized SAM2~\cite{SAM2} to semi-automatically annotate frames from these videos by annotating a few initial frames and utilizing SAM2's tracking capability to complete the annotation of subsequent frames.

We evaluate the real world performance of Pick skill with varying targets in the \{ 3 cubes, 2 plates \} setting. We compare DOCIR with the OCR baseline. We set up the experiments by sampling 30 random initial configurations for the objects as well as target objects for each configuration. We rollout each policy for these 30 episodes and report success rates in~\Cref{tab:real_world_success}. Our DOCIR-based policies are more performant than OCR as also shown earlier in simulation. More importantly, DOCIR shows robust sim2real transfer with almost no drop in performance between simulation and reality compared with OCR. On the other hand OCR policies mostly attempted reaching the right target but failed to correctly grasp.

Furthermore, we highlight that the segmenter implemented in the real world did not always perfectly capture the target object and this could explain some of the performance drop between simulation and reality, yet DOCIR remained reasonably robust to these segmentation errors. We additionally test the robustness of our DOCIR-based skills in varying configurations that include unseen objects in training such as lego blocks and adversarial perturbation during skill execution. DOCIR displays good generalization and robustness in these tests and we provide example policy rollouts in the supplementary video.

\begin{table}[h!]
\centering
\begin{tabular}{lcc}
\toprule
Policy & Success rate (real) & Success rate (sim)\\
\midrule
DOCIR & 0.8 & 0.95\\
OCR & - & 0.49\\
\bottomrule
\end{tabular}
\caption{\textbf{Real world evaluation success rates}. We compare DOCIR's performance on the Pick skill between simulation and the real world. Our results confirm DOCIR is applicable in real-world scenarios and demonstrates good sim2real transfer.}
\label{tab:real_world_success}
\end{table}

\section{Conclusion}
In summary, we have introduced DOCIR, a novel object-centric image representation framework for robotic manipulation based on the core idea of disentangling raw scene images into semantic groups. Throughout our experiments, we demonstrated that our approach improves over state-of-the-art object-centric robotic manipulation both in terms of sample efficiency and final performance. Furthermore, we showed that DOCIR produces policies that are robust to deployment-time changes in the scene.

In future work, we believe there are several avenues that could enlarge the scope of the results obtained in this work. First, while we have demonstrated the feasibility of our approach in the real-world by obtaining a segmentation model fine-tuned for accurately segmenting a predefined set of objects in our scene, we would like to extend this  with a powerful open-world segmentation model. Second, we would like to explore the possible advantages of incorporating large-scale pre-trained models~\cite{R3M, Ma2022VIP, LIV} for processing segmented images. We believe combining both extensions above can lead to faster learning of skills based on a shared disentangled object-centric representation. Third, while we have performed the experiments using reinforcement learning and a simple convolutional network-based policy, our approach can be integrated with other learning paradigms, e.g., imitation learning, and more expressive policies~\cite{chi2023diffusionpolicy, aloha, VQBeT}. Finally, we envision training a high-level policy to leverage the flexibility of DOCIR-based skills by composing them in order to solve long-horizon tasks.

{
    \small
    \bibliographystyle{IEEEtran}
    \bibliography{references}
}

\end{document}